\documentclass{article}

\PassOptionsToPackage{sort}{natbib}
 \usepackage[preprint]{neurips_2025}

\usepackage[utf8]{inputenc} 
\usepackage[T1]{fontenc}    
\usepackage{hyperref}       
\usepackage{url}            
\usepackage{booktabs}       
\usepackage{amsfonts}       
\usepackage{nicefrac}       
\usepackage{microtype}      
\usepackage[dvipsnames]{xcolor}         

\usepackage{graphicx}
\usepackage{subcaption}
\usepackage{dirtytalk}
\usepackage{multirow}
\usepackage{array}
\newcolumntype{?}{!{\vrule width 1pt}}
\usepackage{algorithm}
\usepackage{algpseudocode}
\usepackage{tikz}
\usetikzlibrary{decorations.pathreplacing}
\usetikzlibrary{arrows}
\usepackage{stmaryrd}
\usepackage{enumitem}
\setenumerate{noitemsep,topsep=0pt,parsep=0pt,partopsep=0pt}
\usepackage{wrapfig}

\usepackage{amsmath}
\usepackage{amsthm}
\theoremstyle{definition}
\newtheorem{definition}{Definition}[section]

\theoremstyle{remark}

\newtheorem{proposition}{Proposition}[section]

\title{pySigLib - Fast Signature-Based Computations on CPU and GPU}

\author{%
  Daniil Shmelev\\
  Department of Mathematics\\
  Imperial College London\\
  \texttt{daniil.shmelev23@imperial.ac.uk} \\
  \And
  Cristopher Salvi\\
  Department of Mathematics\\
  Imperial College London\\
  \texttt{c.salvi@imperial.ac.uk} \\
}

\begin{document}

\maketitle

\begin{abstract}
  Signature-based methods have recently gained significant traction in machine learning for sequential data. In particular, signature kernels have emerged as powerful discriminators and training losses for generative models on time-series, notably in quantitative finance. However, existing implementations do not scale to the dataset sizes and sequence lengths encountered in practice. We present \texttt{pySigLib}, a high-performance Python library offering optimised implementations of signatures and signature kernels on CPU and GPU, fully compatible with PyTorch’s automatic differentiation. Beyond an efficient software stack for large-scale signature-based computation, we introduce a novel differentiation scheme for signature kernels that delivers accurate gradients at a fraction of the runtime of existing libraries.
\end{abstract}

\section{Introduction} \label{sec:intro}

Most forms of sequential data such as financial time series, on sufficiently fine time scales, can be represented as a continuous path $x : [0,1] \to \mathbb{R}^d$. It was first shown by \citep{chen1957integration}, and then explored in greater detail and generality in the context of \emph{rough path theory} in \cite{lyons1998differential, hambly2010uniqueness, boedihardjo2016signature}, that any path may be faithfully represented, up to reparameterisation, by the collection of its iterated integrals known as the \emph{path-signature}. The path-signature $S(x)$ is formally defined as the solution to the tensor differential equation \( d y_t = y_t \otimes d x_t \) in the \emph{free tensor algebra} \( T((\mathbb{R}^d)) := \prod_{n=0}^\infty (\mathbb{R}^d)^{\otimes n} \), and thereby can be interpreted as a non-commutative analogue of the exponential function. By applying Picard iteration, one recovers the familiar formulation of the path-signature as the sequence of iterated integrals $S(x) = ( \int_{0<t_1<\cdots<t_n<1} dx_{t_1} \otimes \cdots \otimes dx_{t_n} )_{n \in \mathbb{N}}$. The Stone--Weierstrass theorem ensures that linear functionals acting on the range of the path-signature are dense in the space of continuous real-valued functions defined on compact subsets of \emph{unparameterised paths}. This makes the signature a powerful representation, enabling the approximation of path-dependent functionals using linear models.

Thanks to this \emph{universal approximation property} and a variety of other algebraic features, signature methods have experienced a rapid rise in popularity in recent years, with applications across a wide range of data science domains, including quantitative finance~\cite{perez2020signatures, cuchiero2023signature, abi2025signature, cuchiero2023signature, bayer2025pricing}, cybersecurity~\cite{cochrane2021sk}, information theory~\citep{salvi2021rough, salvi2023structure, shmelev2024sparse}, and even quantum computing~\citep{crew2025quantum}. Signatures have also served as a theoretical foundation for proving universality properties of \emph{neural differential equations}~\cite{morrill2021neural, arribas2020sigsdes} and more recently of \emph{state-space models} (SSMs)~\cite{cirone2024theoretical, cirone2025parallelflow, walker2025structured}. In generative modelling, signatures have enabled new approaches to synthesizing financial time series in a model-independent manner~\citep{buehler2020data}, been used as universal nonlinearities in Seq2Seq architectures~\citep{kidger2019deep}, and provided representation spaces for training \emph{score-based diffusion models} on time series~\citep{barancikovasigdiffusions}. For a detailed and pedagogical introduction to the subject, the reader is referred to~\cite{cass_salvi_notes}, while~\cite{fermanian2023new} offers a survey of recent applications.

In practice, paths are typically obtained via piecewise linear interpolation of discrete time series. Combining a non-trivial algebraic relation known as \emph{Chen's identity} with the elementary fact that the signature of a linear segment is the tensor exponential of its increment, yields a concise expression for the signature of a piecewise linear path \( x = x^1 * \cdots * x^L \) as $S(x) = \exp(\Delta x^1) \otimes \cdots \otimes \exp(\Delta x^L)$. This expression underlies the implementation of signature computations in standard Python libraries such as \texttt{esig}~\cite{esig}, \texttt{iisignature}~\cite{reizenstein2018iisignature}, and \texttt{signatory}~\cite{signatory}. It enables an efficient evaluation of the signature with time complexity \(\mathcal O(Ld^n)\), where \(n \in \mathbb{N}\) is the truncation level. However, due to the exponential growth in the dimension \(d\), the method becomes computationally prohibitive for large \(n\). 

A widely adopted solution to this curse of dimensionality is the use of \emph{signature kernels}~\cite{kiraly2019kernels}. These kernels take the form \(\langle S(x), S(y) \rangle,\) for suitable inner products \(\langle \cdot, \cdot \rangle\) on \(T((\mathbb R^d))\), and can be computed efficiently without explicitly evaluating the feature map \(S(x)\). In particular, recent work has shown that the signature kernel satisfies a Goursat PDE~\citep{salvi2021signature, lemercier2024log}. Signature kernels have since found applications in hypothesis testing~\cite{salvi2021higher, lemercier2021distribution, horvath2023optimal}, causality~\citep{mantensignature}, kernel-based solvers for path-dependent PDEs in derivative pricing under rough volatility~\cite{pannier2024path}, kernel formulations of deep hedging~\cite{cirone2025rough}, and have even been shown to arise as infinite-width limits of neural networks~\cite{cirone2023neural}. They have also been used to train neural SDEs for time-series generation across diverse fields, including fluid dynamics~\cite{salvi2022neural}, computational neuroscience~\cite{holberg2024exact}, and quantitative finance~\cite{issa2023non, diaz2023neural, hoglund2023neural}.
 
Despite these advances, current software implementations of signatures and signature kernels do not scale to large datasets of long time series. Moreover, when signature kernels are employed as loss functions, efficient and accurate backpropagation is crucial. Existing implementations compute derivatives using a second PDE~\citep{lemercier2021siggpde}; while natural, this approach often yields inaccurate gradients, especially for short time series, leading to unreliable model training.

We introduce \texttt{pySigLib}, a high-performance Python library that wraps C++ and CUDA code, offering optimised CPU- and GPU-amenable implementations of signatures and signature kernels, fully compatible with PyTorch’s automatic differentiation. \texttt{pySigLib} is significantly faster than existing packages on both CPU and GPU thanks to algorithmic improvements, optimized memory access, and hardware-level parallelism via SIMD instructions. It also supports efficient on-the-fly application of lead–lag and time-augmentation transformations, providing a substantial speed-up in financial applications. All functions are fully backpropagatable through a PyTorch API. The library is open source and available at \url{https://github.com/daniil-shmelev/pySigLib}
, installable directly via \texttt{pip}, with documentation hosted at \url{https://pysiglib.readthedocs.io}.

\section{Computation of truncated signatures} \label{sec:sig}

We begin with a brief account of the signature transform and numerical methods for computing truncated signatures. For simplicity, we will consider paths of bounded variation. For a detailed introduction to signatures and their applications, we refer the reader to \cite{cass_salvi_notes, fermanian2023new}.

\subsection{The signature transform}

\begin{definition} [Signature Transform]
    Let $\{x_t\}_{a \leq t \leq b}$ be a path of bounded variation. Define the $k^{th}$ \textit{level} of the signature transform as the iterated integral
    \begin{equation*}
    S(x)_{[a,b]}^{(k)} = \int_{a < t_1 < \cdots < t_k < b} dx_{t_1} \otimes dx_{t_2} \otimes \cdots \otimes dx_{t_k} \in V^{\otimes k}
    \end{equation*}
    
    where $\otimes$ is the tensor product and define the signature transform of $x$ to be the formal series of tensors
    \begin{equation*}
        S(x)_{[a,b]} = \left(1, S(x)_{[a,b]}^{(1)}, S(x)_{[a,b]}^{(2)}, S(x)_{[a,b]}^{(3)}, \ldots \right) \in \prod_{i=0}^\infty (\mathbb{R}^d)^{\otimes i}.
    \end{equation*}
    We will often drop the subscript $[a,b]$ when it is clear from context.
\end{definition}

In practical applications, the underlying path $x$ is typically obtained by linear interpolation of a discrete set of data points. The signature of such a path can be computed using the explicit form for the signature of a linear segment and a rule for concatenating signatures, Chen's identity \cite{chen1954iterated}.\par\medskip

\begin{proposition} \label{prop:linearsignature}
    Let $\{x_t\}_{a \leq t \leq b}$ be the linear path joining $x_a$ and $x_b$, $x_t = x_a + \frac{t - a}{b-a}(x_b - x_a)$. Let $z$ denote the increment of the path $x$, viewed as a formal series. That is,
    \begin{equation*}
        z := (1, x_b - x_a, 0, 0, \ldots) \in \prod_{i=0}^\infty (\mathbb{R}^d)^{\otimes i}.
    \end{equation*}
    
    Then the signature of the linear segment is the tensor exponential of the increment,    
    \begin{equation*}
        S(x) = \sum_{k=0}^\infty \frac{z^{\otimes k}}{k!}.
    \end{equation*}
\end{proposition}

\begin{proposition} [Chen's identity { \cite{chen1954iterated}}] \label{prop:chen} Let $\{x_t\}_{a \leq t \leq b}$ and $\{y_t\}_{b \leq t \leq c}$ be paths of bounded variation. Then $S(x * y)_{[a,c]} = S(x)_{[a,b]} \otimes S(y)_{[b,c]}$, where $x*y$ denotes the path concatenation of $x$ and $y$.
\end{proposition}

\subsection{A direct approach}

We implement and test two algorithms for computing truncated signatures. The first of these is the direct approach, as used by \texttt{iisignature} \citep{reizenstein2018iisignature}. Suppose we are given as input a path consisting of points $x_i \in \mathbb{R}^d, 1 \leq i \leq L$, and a truncation level $N$ for the resulting signature. Suppose we have already used the first $\ell$ of these points to construct the signature $S(x_{1 : \ell}) = (A_0, A_1, \ldots, A_N)$. Then the update rule for the next step of the algorithm is given by
\begin{equation*}
    A_k \gets \sum_{i=0}^k A_i \otimes \frac{z^{\otimes (k-i)}}{(k-i)!}
\end{equation*}
where $z := x_{\ell + 1} - x_{\ell}$. The resulting algorithm is outlined in \ref{alg:sig_direct}. Given the computational complexity of signature computations, it is particularly important to minimise costly memory allocation and access operations within the algorithm. Two major design choices allow us to do this: (1) the signature $(A_0, A_1, \ldots, A_N)$ is stored as a single flattened contiguous array; (2) since the update for any given level $A_k$ depends only on $A_i$ for $i < k$, the levels are updated in reverse order, starting at $A_N$ and ending at $A_1$, allowing these to be written directly into the existing memory for $(A_0, A_1, \ldots, A_N)$ without the need for storing intermediary computations.

\begin{algorithm}
\caption{Direct Algorithm for Truncated Signatures}\label{alg:sig_direct}
\textbf{Input:} $\{x_i\}_{1 \leq i \leq L}$, truncation level $N$
\begin{algorithmic}

\State $z = x_1 - x_0$
\State $(A_0, A_1,\ldots, A_N) = (1, z, \ldots, z^{\otimes N} / N!)$ \Comment{See (1) above}

\For {$\ell = 1,\ldots, L-1$}
\State $z = x_{\ell+1} - x_\ell$
\State $(\widetilde{A}_0, \widetilde{A}_1,\ldots, \widetilde{A}_N) = (1, z, \ldots, z^{\otimes N} / N!)$

\For {$k = N, \ldots,1$} \Comment{See (2) above}
\For {$i = k-1,\ldots, 1$} 
\State $A_k = A_k + A_i \otimes \widetilde{A}_{k-i}$
\EndFor

\State $A_k = A_k + \widetilde{A}_k$
\EndFor
\EndFor

\end{algorithmic}
\end{algorithm}

\subsection{Horner's method}

Despite these optimisations, the direct algorithm remains complex, and is largely bound by costly memory access operations. These issues are somewhat improved by Horner's algorithm for polynomial multiplication, as used by \texttt{signatory} \citep{kidger2020signatory}. In Horner's algorithm, the update rule is rewritten to exploit the symmetries of the linear signature, by taking
\begin{align*}
    A_k &= \sum_{i=0}^k A_i \otimes \frac{z^{\otimes(k-i)}}{(k-i)!}\\
    &= \left(\left(\cdots \left(\left(\frac{z}{k} + A_1 \right) \otimes \frac{z}{k-1} + A_2 \right) \otimes \frac{z}{k-2} + \cdots \right) \otimes \frac{z}{2} + A_{k-1} \right) \otimes z + A_k,
\end{align*}
which minimises the number of multiplications necessary, and reduces the number of memory accesses to the right-hand tensor. In our implementation of Horner's algorithm, it will be useful to split the update rule as
\begin{align*}
    A_k &= (B_k + A_{k-1}) \otimes z + A_k\\
    B_k &:= \left(\cdots \left(\left(\frac{z}{k} + A_1 \right) \otimes \frac{z}{k-1} + A_2 \right) \otimes \frac{z}{k-2} + \cdots \right) \otimes \frac{z}{2}
\end{align*}
to allow for additional memory access and allocation optimisations, as we will see below. The resulting algorithm is outlined in \ref{alg:Horner}.

\begin{algorithm}
\caption{Horner's Algorithm for Truncated Signatures}\label{alg:Horner}
\textbf{Input:} $\{x_i\}_{1 \leq i \leq L}$, truncation level $N$
\begin{algorithmic}
\State $z = x_1 - x_0$
\State $(A_0, A_1,\ldots, A_N) = (1, z, \ldots, z^{\otimes N} / N!)$ \Comment{See (1) above}

\For {$\ell = 1,\ldots, L-1$}

\State $z = x_{\ell+1} - x_\ell$

\For {$k = N, \ldots,2$} \Comment{See (2) above}

\State $B_k = z / k$

\For {$i = 1,\ldots, k - 2$}
\State $B_k = B_k + A_i$
\State $B_k = B_k \otimes z/(k-i)$ \Comment{See (3) below}

\EndFor

\State $B_k = B_k + A_{k-1}$
\State $A_k = B_k \otimes z + A_k$ \Comment{See (4) below}

\EndFor
\EndFor

\end{algorithmic}
\end{algorithm}

The two design choices of Algorithm \ref{alg:sig_direct} also apply here. In addition, the following two choices are made: (3) a single, continuous block of memory is pre-allocated to fit $B_N$, and re-used by all intermediate computations involving $B_k$. The multiplication $B_k = B_k \otimes z / (k-i)$ is written directly into the same memory, but carried out in reverse order such that the new values of $B_k$ (of which there are now $d$ times as many) erase the old values only at the very end of the multiplication, when they are no longer required; (4) the last multiplication and addition $B_k \otimes z + A_k$ is written directly into the result, $A_k$.
\begin{center}
    \begin{tikzpicture}[scale=0.7]
    \draw[thick] (0, 2) rectangle (10, 3);
    \draw[thick, fill=RoyalBlue!80] (0, 2) rectangle (2, 3);
    \draw[thick, fill=Maroon!80] (2, 2) rectangle (6, 3);
    \draw[decorate, decoration={brace, amplitude=5pt}, thick] (2, 1.9) -- (0, 1.9) 
        node[midway, below=5pt] {\scriptsize Current $B_k$};

    \draw[decorate, decoration={brace, amplitude=5}, thick] (6, 1) -- (0, 1) 
        node[midway, below=5] {\scriptsize New $B_k = B_k \otimes z / (k-i)$};

    \draw[decorate, decoration={brace, amplitude=5, mirror}, thick] (10, 3.1) -- (0, 3.1) 
        node[midway, above=5pt] {\scriptsize Memory allocated to fit $B_N$};
\end{tikzpicture}
\end{center}

\subsection{Backpropagation through Truncated Signatures} \label{sec:sig_backprop}

Backpropagation is implemented efficiently by using the time-reversed path to deconstruct the signature. We refer the reader to \citep[Section 4.9]{reizenstein2019} for details of the algorithm. Our optimisations of the backpropagation are largely the same as for the forward pass of the signature. A slight modification is made to the original algorithm presented in \citep{reizenstein2019}, whereby Horner's algorithm is used to deconstruct the signature.

\section{Computation of signature kernels} \label{sec:sig_kernel}
We begin with a brief account of signature kernels and numerical techniques to compute them.

\subsection{Signature kernels as the solution of a Goursat PDE}
A key result by \citep{salvi2021signature} established that signature kernels can be computed as the solution of a Goursat-type PDE. Notably, this result states that given two continuous paths $x,y : [0,1] \to \mathbb R^d$, the corresponding signature kernel $k_{x,y}(s,t) := \langle S(x)_s, S(y)_t \rangle$ satisfies the  hyperbolic PDE
\begin{equation*}
    \frac{\partial^2 k_{x,y}}{\partial s \partial t} = \left< \dot{x}_s, \dot{y}_t \right>_V k_{x,y}, \quad k_{x,y}(u,\cdot) = k_{x,y}(\cdot, v) = 1.
\end{equation*}
An effective discretisation of the above PDE is given by \citep{day1966runge, salvi2021signature, wazwaz1993numerical}
\begin{align} \label{eq:kernelFinDiff}
    \widehat{\mathbf{k}}_{i+1, j+1} &= \left(\widehat{\mathbf{k}}_{i+1, j} + \widehat{\mathbf{k}}_{i, j+1}\right) A(\Delta_{i,j}) - \widehat{\mathbf{k}}_{i,j} B(\Delta_{i,j}),\\
    A(\Delta_{i,j}) &= \left(1 + \frac{1}{2} \Delta_{i,j}  + \frac{1}{12}\Delta_{i,j}^2\right), \quad B(\Delta_{i,j}) = \left(1 - \frac{1}{12} \Delta_{i,j}^2\right) \nonumber \\
    \Delta_{i,j} &= \left<x_{t_{i+1}} - x_{t_i}, y_{s_{j+1}} - y_{s_j} \right> \nonumber
\end{align}

over the dyadically refined grid $P_{\lambda_1, \lambda_2} = \{(t_i, s_j) : 0 \leq i \leq 2^{\lambda_1} L_1, \hspace{1mm} 0\leq j \leq 2^{\lambda_2} L_2\}$ of order $(\lambda_1, \lambda_2)$, where $L_1, L_2$ are the lengths of the discrete data streams $x,y$ respectively.

\subsection{An algorithm for signature kernels}

Implementations of signature kernels exist for both CPU and GPU, most notably via the \texttt{sigkernel} and \texttt{sigkerax} packages \citep{salvi2021signature}. The resulting algorithm is outlined in \ref{alg:sigkernel_cpu}. In our implementation, several choices are notable: (1) as opposed to other packages, we allow $\lambda_1 \neq \lambda_2$, which can prove useful when $x$ and $y$ are of significantly different lengths; (2) all values of $\Delta_{i,j}$ required for \eqref{eq:kernelFinDiff} are precomputed using a \texttt{matmul} operation. In cases where the dimension of the underlying path is large, this operation accounts for almost all of the algorithm’s runtime, so it is critical that it is highly optimized. In \texttt{pySigLib}, this is realised using PyTorch’s \texttt{bmm} function \citep{paszke2019pytorch}; and (3) whereas other packages implementing signature kernels precompute the dyadically refined path, \texttt{pySigLib} applies the dyadic refinement on-the-fly, improving both computational and memory efficiency.

\begin{algorithm}
\caption{CPU Algorithm for Signature Kernels}\label{alg:sigkernel_cpu}
\textbf{Input:} $\{x_i\}_{1 \leq i \leq L_1}, \{y_i\}_{1 \leq i \leq L_2}, \lambda_1, \lambda_2$ \Comment{See (1) above}
\begin{algorithmic}
\State $dx = x_{1:L_1} - x_{0:L_1 - 1}; \quad dy = y_{1:L_1} - y_{0:L_1 - 1}$
\par\smallskip
\State $\Delta = \texttt{matmul}(dx^T, dy)$ \Comment{See (2) above}
\par\smallskip
\For {$s = 0, \ldots, 2^{\lambda_1} L_1 - 1; \quad t = 0, \ldots, 2^{\lambda_2} L_2 - 1$}
    \State $p = \Delta[s \sslash 2^{\lambda_1}, t \sslash 2^{\lambda_1}]$ \Comment{See (3) above}
    \State $\widehat{\mathbf{k}}[s,t] = (\widehat{\mathbf{k}}[s-1, t] + \widehat{\mathbf{k}}[s,t-1])* (1 + \frac{1}{2}p + p^2) - \widehat{\mathbf{k}}[s-1, t-1] * (1 - p^2)$
\EndFor
\State \Return $\widehat{\mathbf{k}}[-1, -1]$

\end{algorithmic}
\end{algorithm}

\subsection{Exploiting GPU parallelism}

\begin{wrapfigure}{r}{6.2cm}
    \vspace{-0.6cm}
    \begin{tikzpicture}[scale=0.18] 
        \foreach \x in {0,...,29} {
            \fill[ForestGreen!80] (\x, 20) rectangle ++(1,-1);
        }

        \foreach \x in {0,...,29} {
        
            \fill[ForestGreen!80] (\x, 12) rectangle ++(1,-1);
        }

        \foreach \x in {0,...,29} {
            \fill[ForestGreen!80] (\x, 4) rectangle ++(1,-1);
        }

        \foreach \x in {12,...,19} {
            \fill[Maroon!80] (\x-2, \x) rectangle ++(1,-1);
            \fill[Maroon!80] (\x-1, \x) rectangle ++(1,-1);
            \fill[Maroon!80] (\x, \x) rectangle ++(1,-1);
        }

        \draw[step=1cm,gray,thin] (0,0) grid (30,20);

        \draw[decorate,decoration={brace,amplitude=5pt}, thick] (-0.5,11) -- (-0.5,19);
        \node[anchor=east, rotate=90] at (-2.5,18.5) {\scriptsize Block of 32};

        \begin{scope}[shift={(9,-2)}] 
            \fill[ForestGreen!80] (0,0) rectangle ++(1,1);
            \node[right] at (1,0.4) {\scriptsize Global};

            \fill[Maroon!80] (7,0) rectangle ++(1,1);
            \node[right] at (8,0.4) {\scriptsize Shared};
        \end{scope}

    \end{tikzpicture}
    \vspace{-0.5cm}
\end{wrapfigure}

As suggested in \citep{salvi2021signature}, the algorithm can be effectively parallelised on a GPU, by noting that elements on any given anti-diagonal of the PDE grid exhibit no interdependencies, and so can be computed in parallel. Following this approach, we note that it is not necessary to store the entire PDE grid. Rather, one can store only the current anti-diagonal and the two before it. As the algorithm progresses, the three anti-diagonals are \say{rotated}, so that the first anti-diagonal becomes the second, the second becomes the third, and the third is overwritten with new values to become the first. This dramatically reduces the required memory allocation, allowing for the three anti-diagonals to be kept in shared memory of the GPU, which has a significantly lower latency than the global memory which a full PDE grid would be forced to reside in.

As of writing, when using other signature kernel libraries for GPU computations, the length of the input streams is severely limited by the thread count of the GPU, as the algorithm attempts to assign one thread per entry of the diagonal. We avoid this problem by computing kernels in blocks of $32$. For the first such block, a vector of initial conditions in global memory is populated with all ones, acting as the first row of the PDE grid. As the block progresses, the initial condition is overwritten by the final row in the block, which becomes the initial condition for the second block of $32$. As such, one is never limited by the thread count, as only $32$ threads are allocated per kernel computation. Naturally, this means the GPU is underutilized when computing a single kernel. However, when a batch of kernels are computed, blocks associated with different kernels are allowed to run asynchronously with respect to each other, meaning all available threads are used when the batch size is sufficiently large. We anticipate this should be the case in almost all practical applications.

\subsection{Backpropagation through signature kernels} \label{sec:sig_kernel_backprop}

Existing methods for backpropagation approximate derivatives as the solution to another PDE \citep{lemercier2021siggpde}. The downside to this is that the gradients are not exact. When either the length of the underlying path or the dyadic order is too low, this difference is significant. To avoid this issue, we compute exact gradients by differentiating through the operations of the solver \eqref{eq:kernelFinDiff}. As we will see, the resulting algorithm is significantly faster than the PDE approach, whilst producing exact gradients. Given a scalar function $F$ and its derivative with respect to the signature kernel $\partial F / \partial \widehat{\mathbf{k}}_{L_1,L_2}$, we aim to compute $\partial F / \partial \Delta_{a,b}$ for all indices $a,b$, which can then easily be used to compute the derivatives with respect to the underlying paths $x$ and $y$ via a linear transformation. A naive approach may be to try and differentiate the above relation directly with respect to $\Delta_{a,b}$ to try and form an update rule for $\partial \widehat{\mathbf{k}}_{i,j} / \partial \Delta_{a,b}$. This approach would require traversing a PDE grid for each pair $a,b$, resulting in a serial complexity of $\mathcal{O}(2^{\lambda_1 + \lambda_2} L_1^2 L_2^2 )$. One can, however, achieve the same result in only one traversal of a PDE grid, by targeting instead $\partial F / \partial \widehat{\mathbf{k}}_{a,b}$. For simplicity, assume our grid is not dyadically refined, i.e. $\lambda_1 = \lambda_2 = 0$. For convenience, define $\partial F / \partial \widehat{\mathbf{k}}_{i,j} = 0$ if $i > L_1$ or $j > L_2$. Then we have the following relations, computable in one backward pass of a PDE grid:
\begin{align*}
    \frac{\partial F}{\partial \widehat{\mathbf{k}}_{i,j}} & = \frac{\partial F}{\partial \widehat{\mathbf{k}}_{i+1,j}} A(\Delta_{i,j-1}) + \frac{\partial F}{\partial \widehat{\mathbf{k}}_{i,j+1}} A(\Delta_{i-1,j}) - \frac{\partial F}{\partial \widehat{\mathbf{k}}_{i+1,j+1}} B(\Delta_{i,j}),\\
    \frac{\partial F}{\partial \Delta_{i,j}} &= \frac{\partial F}{\partial \widehat{\mathbf{k}}_{i+1,j+1}} \left[ \left(\widehat{\mathbf{k}}_{i+1,j} + \widehat{\mathbf{k}}_{i,j+1}\right) \left(\frac{1}{2} + \frac{1}{6} \Delta_{i,j}\right) + \frac{1}{6} \widehat{\mathbf{k}}_{i,j} \Delta_{i,j} \right].
\end{align*}

The same approach applies to dyadically refined grids, noting that in this case there are multiple indices $a,b$ for which $\widehat{\mathbf{k}}_{a,b}$ depends on $\Delta_{i,j}$.

\begin{algorithm}
\caption{CPU Algorithm for Backpropagation through Signature Kernels}\label{alg:sigkernelbackprop_cpu}
\textbf{Input:} $x_i \in \mathbb{R}^d$, $1 \leq i \leq L_1$ and $y_i \in \mathbb{R}^d$, $1 \leq i \leq L_2$\\
\textbf{Input:} Dyadic order for $x$, $\lambda_1$, and dyadic order for $y$, $\lambda_2$\\
\textbf{Input:} Dyadically refined kernel grid $\widehat{\mathbf{k}}$, precomputed or computed on-the-fly.\\
\textbf{Input:} Derivative $\partial F/\partial\widehat{\mathbf{k}}[s,t]$.
\begin{algorithmic}
\State $dx = x_{1:L_1} - x_{0:L_1 - 1}, \quad dy = y_{1:L_1} - y_{0:L_1 - 1}; \quad \Delta = \texttt{matmul}(dx^T, dy)$
\par\smallskip
\State $\Delta = \texttt{matmul}(dx^T, dy)$
\par\smallskip

\For {$s = 2^{\lambda_1} L_1 - 1, \ldots, 0; \quad t = 2^{\lambda_2} L_2 - 1, \ldots, 0$}
    \State $p_{i,j} = \Delta[i \sslash 2^{\lambda_1}, j \sslash 2^{\lambda_2}]$ \quad for $(i,j) = (s,t), (s-1, t), (s, t-1)$

    \State $d_1[s,t] = (d_1[s+1, t] * A(p_{s,t-1}) + d_1[s, t+1] * A(p_{s-1,t}) - d_1[s+1, t+1] * B(p_{s,t}))$
    \State $d_2[s \sslash 2^{\lambda_1}, t \sslash 2^{\lambda_2}] \mathrel{+}= d_1[s,t] * \{(\widehat{\mathbf{k}}[s+1,t] + \widehat{\mathbf{k}}[s,t+1]) * A'(p_{s,t}) - \widehat{\mathbf{k}}[s,t] * B'(p_{s,t})\}$
\EndFor
\State \Return $d_2$ \Comment{Result is $\partial F / \partial \Delta$, which can be transformed into $\partial F / \partial x$ or $\partial F / \partial y$}

\end{algorithmic}
\end{algorithm}

\section{Path Transformations} \label{sec:path_transforms}

Given a path, it is common to apply certain path-to-path transformations before computing a transform such as the signature. The most common path-to-path transformations are time-augmentation and the lead-lag transform. pySigLib provides backpropagatable implementations of these transforms. For signature computations, pySigLib is capable of adapting the algorithms internally to perform these transformations on-the-fly, which is often faster and more memory-efficient than pre-computing them. Given a path consisting of points $x_{t_i} \in \mathbb{R}^d$, the time augmented path is defined by $\hat{x}_{t_i} := (x_{t_i}, t_i) \in \mathbb{R}^{d+1}$. Time-augmentation is useful for introducing time-reparametrisation variance to the signature, in applications where the timing of events is critical. The lead-lag transformation is defined by $X^{LL}_{t_i} := (X^{\text{Lead}}_{t_i}, X^{\text{Lag}}_{t_i})$, where \citep{chevyrev2016primer, gyurko2013extracting, flint2016discretely}
\begin{align*}
    X^{\text{Lead}}_{t_i} &:= \begin{cases}
            X_{t_k} & \text{if } i = 2k, \\
            X_{t_k} & \text{if } i = 2k - 1,
        \end{cases}\\
    X^{\text{Lag}}_{t_i} &:= \begin{cases}
            X_{t_k} & \text{if } i = 2k, \\
            X_{t_k} & \text{if } i = 2k + 1.
        \end{cases}
\end{align*}

The lead-lag transform is particularly useful for feature-extraction applied to financial data streams, where the signature of the lead-lag path may be used as an approximation to the It\^o-signature \citep{flint2016discretely, gyurko2013extracting}.
    
\section{Performance} \label{sec:performance}

A 14-core i7-13700H CPU with 32GB of RAM and an NVIDIA RTX 4060 GPU are used with Windows 11 24H2 and Python 3.9. For all experiments, the minimum runtime is taken over 50 runs. CUDA Toolkit 11.8 is used. \texttt{pySigLib} version 0.2.0 is compiled using MSVC toolset version 14.44.

\subsection{Truncated signatures} \label{sec:sig_performance}

\begin{figure*}[h!]
    \centering
    \begin{subfigure}[t]{0.45\textwidth}
        \centering
        \includegraphics[width = \textwidth]{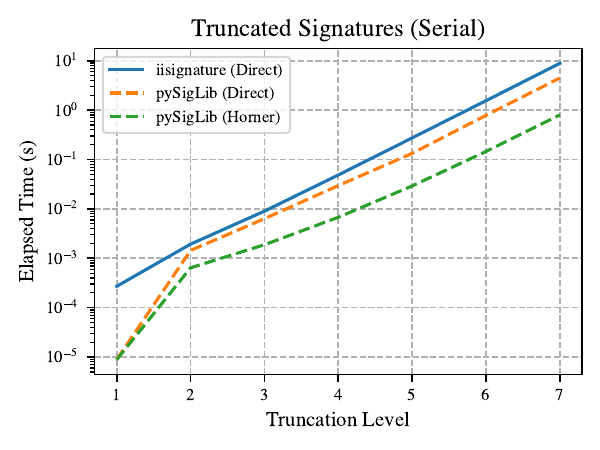}
    \end{subfigure}%
    ~ 
    \begin{subfigure}[t]{0.45\textwidth}
        \centering
        \includegraphics[width = \textwidth]{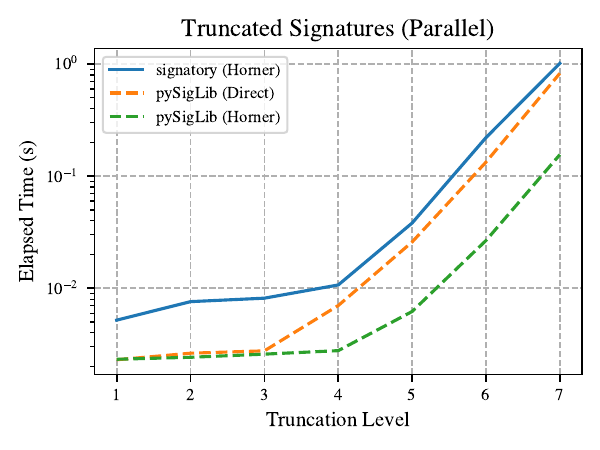}
    \end{subfigure}
    \caption{\textit{Timings for a batch of $32$ paths of length $1024$, dimension $5$.}}
    \label{fig:sig_runtime}
\end{figure*}

Table \ref{table:sig_runtime} and Figure \ref{fig:sig_runtime} show the runtime for signatures and their backpropagation. Runtimes are compared to \texttt{esig} \cite{esig}, version 1.0.0, \texttt{iisignature} \cite{reizenstein2018iisignature}, version 0.24 and \texttt{signatory} \cite{signatory}, version 1.2.6.

\begin{table}[h!]
\hspace*{-0.9cm}
\centering
\scalebox{0.8}{
\begin{tabular}{c?ccc|cc}
\multirow{2}{*}{(B, L, d, N)} &
  \multicolumn{3}{c|}{Forward (CPU, serial)} &
  \multicolumn{2}{c}{Forward (CPU, parallel)} \\
\cline{2-6}
& esig & iisignature & pySigLib & signatory & pySigLib \\
\hline
(128, 256, 4, 6) & 1.1310 & 0.4104 & \textbf{0.0482} & 0.0558 & \textbf{0.0110} \\
(128, 512, 8, 5) & 11.4814 & 4.7908 & \textbf{0.3673} & 0.4512 & \textbf{0.0896} \\
(128, 1024, 16, 4) & 34.7836 & 14.3296 & \textbf{1.1512} & 2.2121 & \textbf{0.2988} \\
\cline{1-6}
\multirow{2}{*}{(B, L, d, N)} &
  \multicolumn{3}{c|}{Backward (CPU, serial)} &
  \multicolumn{2}{c}{Backward (CPU, parallel)} \\
\cline{2-6}
& esig & iisignature* & pySigLib & signatory & pySigLib \\
\hline
(128, 256, 4, 6) & N/A & 2.3360 & \textbf{0.7456} & 0.5918 & \textbf{0.1212} \\
(128, 512, 8, 5) & N/A & 28.9140 & \textbf{8.7427} & 5.9214 & \textbf{1.6595} \\
(128, 1024, 16, 4) & N/A & 87.2166 & \textbf{27.5991} & 18.2688 & \textbf{5.5374} \\
\end{tabular}}
\caption{\textit{Runtime (seconds) for a batch of $B$ paths of length $L$, dimension $d$ and a truncation level of $N$. *\texttt{iisignature} recomputes the signature during the backward pass, and this is included in the runtime.}}
\label{table:sig_runtime}
\end{table}

\subsection{Signature kernels} \label{sec:sig_kernel_performance}

\begin{figure*}[h!]
    \centering
    \begin{subfigure}[t]{0.45\textwidth}
        \centering
        \includegraphics[width = \textwidth,trim={.3cm .3cm .3cm .3cm},clip]{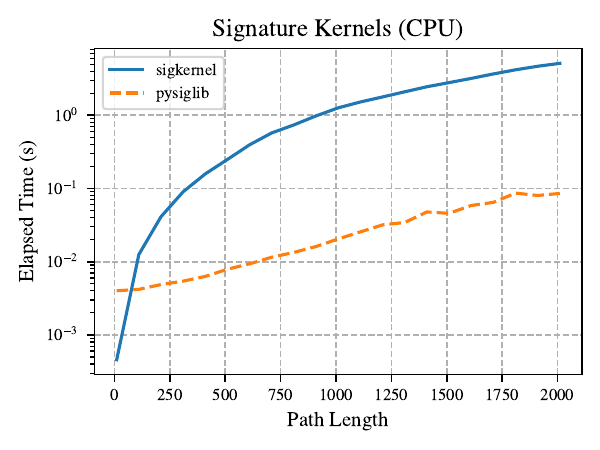}
    \end{subfigure}%
    ~ 
    \begin{subfigure}[t]{0.45\textwidth}
        \centering
        \includegraphics[width = \textwidth,trim={.3cm .3cm .3cm .3cm},clip]{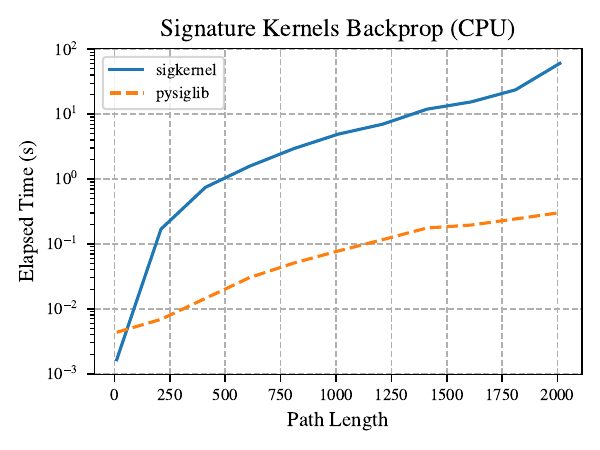}
    \end{subfigure}
    \begin{subfigure}[t]{0.45\textwidth}
        \centering
        \includegraphics[width = \textwidth,trim={.3cm .3cm .3cm .3cm},clip]{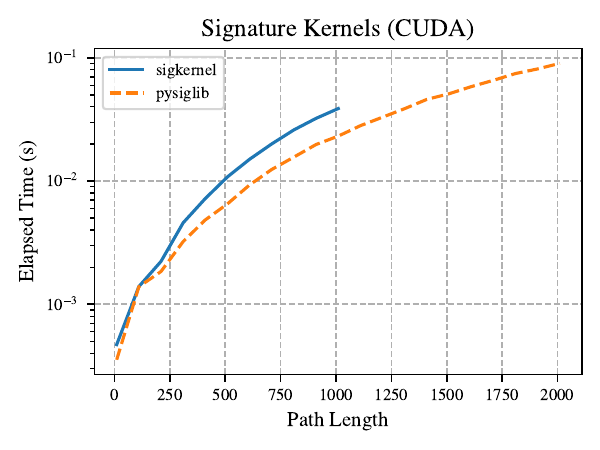}
    \end{subfigure}%
    ~ 
    \begin{subfigure}[t]{0.45\textwidth}
        \centering
        \includegraphics[width = \textwidth,trim={.3cm .3cm .3cm .3cm},clip]{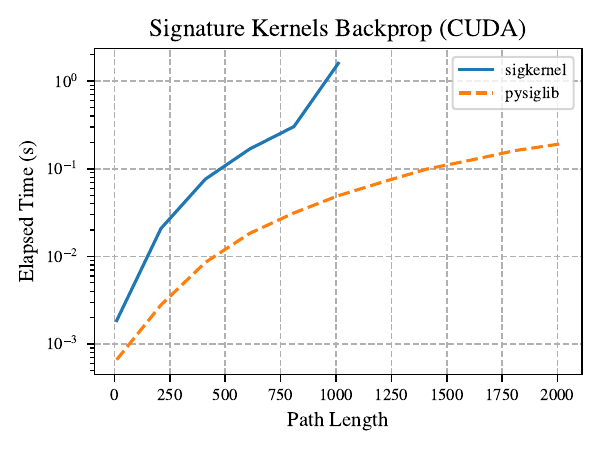}
    \end{subfigure}
    \caption{\textit{Runtime (seconds) for signature kernels and backpropagation, computed for a batch of 32 paths of dimension 5.}}
    \label{fig:runtime_kernel}
\end{figure*}

Table \ref{table:runtime_kernel} and Figure \ref{fig:runtime_kernel} show the runtime of signature kernels and their backpropagation. Runtimes are compared to the \texttt{sigkernel} package. \texttt{sigkernel} is limited by the thread count (1024) on GPU and memory on CPU. In Table \ref{table:runtime_kernel}, a dash indicates that these issues prevented the computation from running.

\begin{table}[h!]
\hspace*{-0.9cm}
    \centering
    \scalebox{0.8}{
    \begin{tabular}{c?cc|cc|cc|cc}
    \multirow{2}{*}{(B, L, d)} &
      \multicolumn{2}{c|}{Forward (CPU)} &
      \multicolumn{2}{c|}{Forward (GPU)}  & 
      \multicolumn{2}{c|}{Backward (CPU)} &
      \multicolumn{2}{c}{Backward (GPU)}\\
    \cline{2-9}
    & sigkernel & pySigLib & sigkernel & pySigLib & sigkernel & pySigLib & sigkernel & pySigLib\\
    \hline
    (128, 256, 8) & 0.3610 & \textbf{0.0118} & 0.0088 & \textbf{0.0034} & 2.3248 & \textbf{0.0322} & 0.1475 & \textbf{0.0057}\\
    (128, 512, 16) & 1.3608 & \textbf{0.0364} & 0.0376 & \textbf{0.0117} & 15.0445 & \textbf{0.1276} & 15.3843 & \textbf{0.0231}\\
    (128, 1024, 32) & 6.0874 & \textbf{0.2325} & - & \textbf{0.0653} & - & \textbf{0.6019} & - & \textbf{0.1112}
    \end{tabular}}
    \caption{\textit{Runtime (seconds) for a batch of $B$ paths of length $L$ and dimension $d$. Dyadic order is set to 0. A dash indicates the computation failed.}}
    \label{table:runtime_kernel}
\end{table}

\section{Conclusion}

We have introduced \texttt{pySigLib}, a computational library designed to overcome the challenges of applying signature-based methods to large-scale data. Our results show significant improvements in the computation of both signatures and signature kernels, allowing for efficient processing of long time-series data. We presented a novel approach to backpropagation through signature kernels, which achieves accurate gradients at a fraction of the runtime of existing implementations. Concrete experiments integrating \texttt{pySigLib} into deep learning pipelines are beyond the scope of this paper and are left for future work.

\bibliographystyle{abbrv}
\bibliography{references}



\end{document}